# AI-Driven Smartphone Solution for Digitizing Rapid Diagnostic Test Kits and Enhancing Accessibility for the Visually Impaired

R. B. Dastagir[1], J. T. Jami[1], S. Chanda[1], F. Hafiz[2,3], M. Rahman[4], K. Dey[5], M. M. Rahman[6], M. Qureshi[7] and M. M. Chowdhury[8]

[1]Department of Mechanical Engineering, Bangladesh University of Engineering and Technology, Dhaka-1000, Bangladesh
[2]Department of Electrical and Electronic Engineering, Bangladesh University of Engineering and Technology, Dhaka-1000, Bangladesh
[3]Department of Computer Science and Engineering, United International University, Dhaka-1212, Bangladesh
[4]Department of Electrical and Computer Engineering, North Carolina State University, Raleigh, NC 27695, United States
[5]Department of Metallurgical and Materials Engineering, University of Alabama, Tuscaloosa, AL 35487, United States
[6]School of Electrical and Computer Engineering, Purdue University, West Lafayette, Indiana 47907, United States
[7]LifeScape Bioscience Inc., Vancouver, British Columbia V6B 1H4, Canada
[8]Department of Electrical and Computer Engineering, University of British Columbia, Vancouver, British Columbia V6T 1Z4, Canada

## Abstract

Rapid diagnostic tests are crucial for timely disease detection and management, yet accurate interpretation of test results remains challenging. In this study, we propose a novel approach to enhance the accuracy and reliability of rapid diagnostic test result interpretation by integrating artificial intelligence (AI) algorithms, including convolutional neural networks (CNN), within a smartphone-based application. The app enables users to take pictures of their test kits, which YOLOv8 then processes to precisely crop and extract the membrane region, even if the test kit is not centered in the frame or is positioned at the very edge of the image. This capability offers greater accessibility, allowing even visually impaired individuals to capture test images without needing perfect alignment, thus promoting user independence and inclusivity. The extracted image is analyzed by an additional CNN classifier that determines if the results are positive, negative, or invalid, providing users with the results and a confidence level. Through validation experiments with commonly used rapid test kits across various diagnostic applications, our results demonstrate that the synergistic integration of AI significantly improves sensitivity and specificity in test result interpretation. This improvement can be attributed to the extraction of the membrane zones from the test kit images using the state-of-the-art YOLO algorithm. Additionally, we performed SHapley Additive exPlanations (SHAP) analysis to investigate the factors influencing the model's decisions, identifying reasons behind both correct and incorrect classifications. By facilitating the differentiation of genuine test lines from background noise and providing valuable insights into test line intensity and uniformity, our approach offers a robust solution to challenges in rapid test interpretation. Overall, our findings suggest that integrating AI-powered analysis represents a promising advancement in point-of-care diagnostics, with potential to enhance accessibility for visually impaired individuals and applications in healthcare, veterinary medicine, and environmental monitoring.

## 1. Introduction

Rapid diagnostic tests are essential for promptly detecting and managing various diseases and health conditions. They provide quick results at the point of care, facilitating immediate decision-making and treatment initiation. Despite their benefits, accurately interpreting rapid test results, especially when test lines are faint or ambiguous, presents a significant challenge for both healthcare providers and users. Misinterpreting results can result in delayed diagnosis, inappropriate treatment decisions, and potentially adverse health outcomes. Rapid diagnostic kits often work based on lateral flow assays (LFAs). LFAs have evolved significantly from their initial use in pregnancy testing to becoming widely employed in infectious disease diagnostics. LFAs have evolved significantly over the years, as evidenced by key studies. Posthuma-Trumpie et al. [1] contributed to optimizing LFAs by enhancing sensitivity, specificity, and usability. Koczula and Gallotta [2] provided a comprehensive review, discussing the principles, applications, and challenges of LFAs, highlighting improvements in assay performance. Mattioli et al. [3] raised concerns about potential interpretation issues due to human eye variability. Saisin et al. [5] noted advancements in sensitivity and accuracy with camera-based lateral flow immunoassay readers. Long et al. [6] highlighted the rapid development of SARS-CoV-2 serological rapid diagnostic tests (RDTs), emphasizing LFAs' role in enabling widespread testing during global health crises. Mendels et al. [7] and van Grinsven et al. [4] developed and evaluated ALFA, an automated system using machine learning for accurate analysis of LFIA self-test results. García-Fiñana et al. [8] analyzed city-wide testing programs using LFAs and compared cycle threshold (Ct) levels from follow-up PCR tests. Peto [9] provided insights into the sensitivity and specificity of LFAs in mass testing scenarios. Such advancements are pivotal in expanding the utility of LFAs beyond clinical laboratories to resource-limited environments, aiding in early detection and containment of infectious outbreaks.

RDTs have evolved significantly in recent years, becoming pivotal tools in detecting infectious diseases. Wang et al. [10] highlighted the expansive applications of RDTs, underscoring their versatility across various infectious diseases through lateral flow immunoassays (LFIA). Foundation for Innovative New Diagnostics [11] provided an overview of the landscape of COVID-19 antibody RDTs, emphasizing their diverse applications and potential utility in clinical settings. Liu et al. [12] addressed challenges in RDT interpretation due to varying antibody levels, crucial for optimizing diagnostic accuracy. Lastly, Dortet et al. [13] evaluated the xRCovid app's performance across multiple SARS-CoV-2 serological RDTs, demonstrating enhanced diagnostic yield and reliability. Lee et al. [14] discussed the efficacy of rapid antigen tests (RATs) in detecting culturable virus compared to qPCR, illustrating ongoing advancements in RDT technologies. The development of serological RDTs by Gonzalez et al. [15] enables the assessment of population-wide immunity and vaccine effectiveness against emerging pathogens. These tests detect antibodies produced in response to infection, providing insights into disease prevalence and transmission dynamics within communities. Advances in RDT technology continue to focus on improving assay sensitivity, specificity, and usability, critical for timely diagnosis and public health interventions during outbreaks.

The integration of smartphone applications has democratized access to diagnostic capabilities, transforming how individuals and healthcare providers interact with diagnostic tools. In recent years, smartphone applications have significantly advanced in their capabilities for healthcare and diagnostic purposes. Beginning with Sim et al. [16], their work explored the integration of smartphone apps equipped with detection aids into clinical settings, demonstrating their potential to enhance diagnostic accuracy and accessibility. Building upon this, Gan & Poon [17] further elucidated how these apps can replace traditional scientific equipment in certain clinical scenarios, paving the way for cost-effective and portable diagnostic solutions. Moving forward, Cheong et al. [18] contributed to the ongoing evolution by expanding our understanding of smartphone app functionalities in medical contexts, emphasizing their role in facilitating remote monitoring and real-time diagnostics. More recently, Gan & Yeo [19] likely continued this trajectory with advancements that have broadened the scope of smartphone app technologies applicable to healthcare, potentially integrating artificial intelligence and machine learning for enhanced diagnostic capabilities. These studies collectively underscore the transformative potential of smartphone applications in revolutionizing diagnostic technologies, promising increased efficiency and accessibility in healthcare delivery.

Artificial intelligence (AI) has emerged as a powerful tool in diagnostic medicine, revolutionizing the accuracy and efficiency of disease detection. Krizhevsky et al. [20] developed the xRCovid app, employing an artificial neural network (ANN) for standardized interpretation of rapid diagnostic test (RDT) results, laying foundational work in automated diagnostic tools. Joung et al. [21] explored digitalized image analysis of LFAs, aiming to enhance diagnostic accuracy and reliability through advanced digital processing techniques. Zeng et al. [22] investigated convolutional neural networks (CNNs) for diagnostic purposes, focusing on improving detection sensitivity and specificity, showcasing the potential of AI in medical diagnostics. Budd et al. [23] introduced supervised contrastive learning for adjusting weights in pre-trained models using mixed data, expanding the robustness of model adaptation. Mendels et al. [24] likely contributed to AI-based detection of SARS-CoV-2 antibodies, leveraging deep learning techniques to enhance diagnostic accuracy amid the pandemic. Jing et al. [25] introduced the AutoAdapt POC approach, integrating automated membrane extraction, self-supervised learning, and few-shot learning to efficiently adapt pre-trained models to new point-of-care diagnostic tests, illustrating innovative advancements in diagnostic technology. Ward et al. [26] and Dai [27] highlighted machine learning systems such as ALFA in improving population-level immunity evaluation through serological testing for SARS-CoV-2, showcasing the pivotal role of AI in public health assessments. Machine learning algorithms such as CNNs and recurrent neural networks (RNNs) are employed to detect subtle patterns in diagnostic images or data streams, surpassing human capabilities in speed and accuracy. The application of AI in diagnostic tools not only improves diagnostic precision but also facilitates early detection and intervention, crucial for mitigating the spread of infectious diseases like COVID-19.

Despite remarkable progress, challenges persist in the widespread adoption and standardization of these technologies. Ibitoye et al. [28] highlighted challenges associated with LFAs, particularly focusing on operational errors and result interpretation issues that impact their reliability in decentralized settings such as primary care clinics and homes. Building on this, Carrio et al. [29] discussed quality control challenges specific to LFAs, emphasizing the necessity for standardized processes to ensure consistent and reliable diagnostic outcomes across various settings and users. Moving forward to Corman et al. [30] highlighted issues such as assay variability, regulatory complexities, and the need for robust validation protocols to ensure diagnostic reliability across diverse settings. In the same year, Andryukov [31] contributed to the field by reviewing advancements in diagnostic technologies, including LFAs, within the context of point-of-care diagnostics. Recently, Atchison et al. [32] addressed uncertainties and challenges in deploying SARS-CoV-2 antigen LFAs, highlighting factors influencing test performance and their implications for public health monitoring. Additionally, Flower et al [33] developed an LFIA used in the REACT-2 study, which reported specific sensitivity and specificity values, further contributing to the ongoing refinement and application of LFAs in diagnostic practices.

In this study, we introduce a novel approach to improving the accuracy and reliability of rapid test result interpretation by integrating smartphone-based AI, with a particular focus on enhancing accessibility for visually impaired users. Our approach leverages advanced AI algorithms, including Convolutional Neural Networks (CNNs) and YOLOv8, to digitize test images captured by smartphone cameras and analyze test lines.

The primary goal of our research is to overcome the limitations of conventional rapid test interpretation methods, providing a robust solution that improves diagnostic accuracy and efficiency. We focus on developing a custom AI model for rapid test result detection, benchmarking its performance against existing models to assess its capabilities. Another key objective is to identify and understand the causes of misclassifications by investigating errors through techniques such as SHAP analysis. Additionally, we aim to utilize smartphone-aided AI to drive significant advancements in point-of-care diagnostics, ensuring that our solution is accessible and reliable for all users, including those with visual impairments.

## 2. Methodology

In this study, we employed a comprehensive methodology to develop an automated system for analyzing COVID-19 rapid test kits using deep learning. The process began with the collection and preparation of a diverse dataset comprising images from a previous study and newly captured photographs. These images were manually annotated to identify the membrane regions of the test kits. We then trained a YOLO v8 neural network to detect and extract these membrane zones accurately.

Subsequently, the extracted regions were utilized to train another Convolutional Neural Network (CNN) to classify the test results as positive, negative, or invalid. Finally, we developed a user-friendly mobile application, enabling users to capture images of rapid test kits and receive immediate and accurate test result analysis.

### 2.1 Dataset Collection and Preparation

The dataset for training the neural network included images from two sources: an existing prepared dataset and newly captured photographs of COVID-19 rapid test kits. The existing dataset, labeled "Positive" and "Negative" [24], is shown in Figure 1. We also collected new images, including positive, negative, and invalid test kits, using phone cameras under various lighting conditions and angles, expanding the classes to three as shown in Figure 2. This enhanced the model's robustness and realism for rapid test detection. The new dataset comprises 1015 positive, 1389 negative, and 164 invalid images. The older dataset, described in Table 1, contains images of antibody-detecting RDTs captured from the same elevation and illumination. Each image was reviewed at least twice by human analysts, with a third review if needed. Additionally, we created a novel dataset variation under different lighting and background conditions using a high-resolution camera at various angles and orientations, as shown in Figure 3.

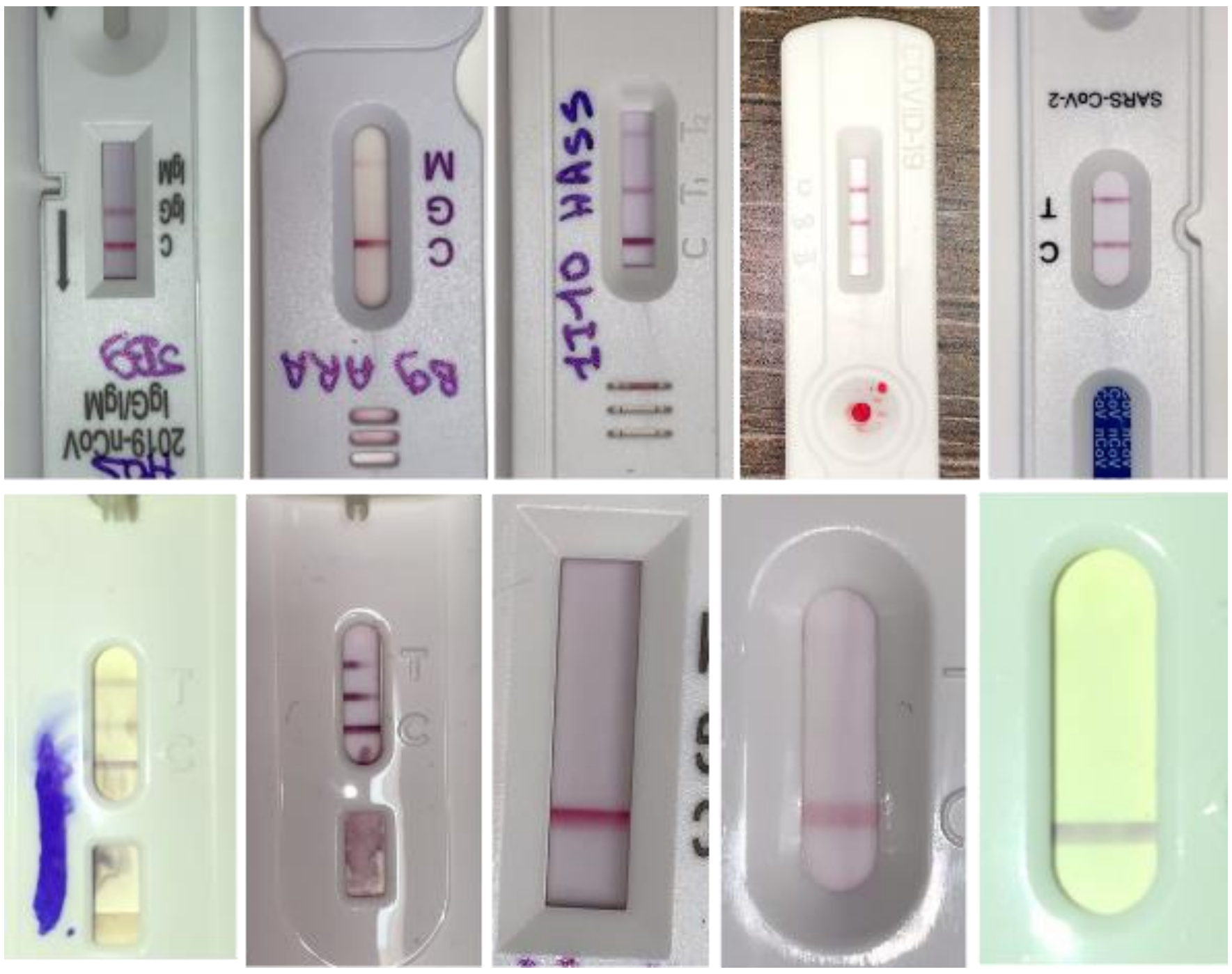

**Figure 1:** Mendels et al. [24] prepared an image dataset of rapid test kits with positive and negative results confirmed by two human analysts. The images were taken using a camera positioned at an elevation of 98 mm from the reference plane. Selective images from this dataset were used in the current work to train and validate the performance of our model. Images of rapid test kits from multiple brands with “POSITIVE” and “NEGATIVE” results are presented in this Figure.

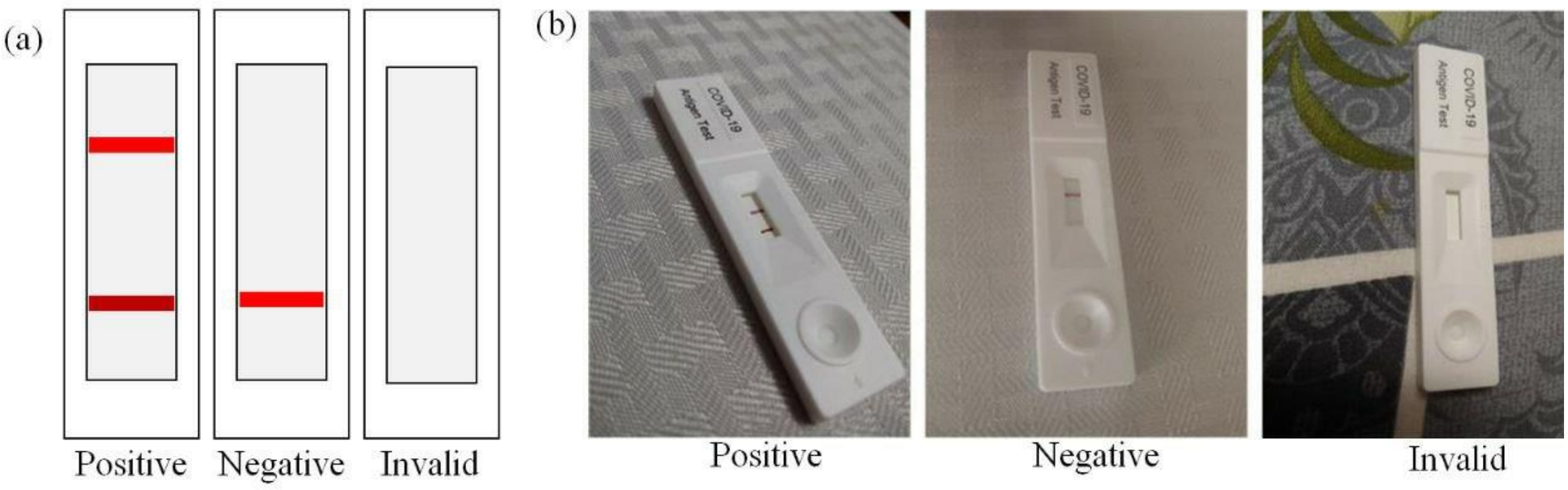

**Figure 2:** The dataset was enhanced in the current study with the addition of a third class, the "INVALID" class typically seen with the absence of both the control line and test line. Images taken in our own setup at different elevations and angles of the three classes are merged in the dataset. (a) Schematic diagram of the three classes, "POSITIVE", "NEGATIVE" and "INVALID" (b) Images were taken at our own setup showing all three results, "POSITIVE", "NEGATIVE" and "INVALID" respectively.

**Table 1:** List of Rapid Diagnostic Tests (RDTs) for COVID-19, including their Manufacturers and Locations [13]

| RDT Name | Manufacturer | Location |
|---|---|---|
| NG-Test IgG-IgM COVID-19 | NG-Biotech | Guipry, France |
| Anti-SARS-CoV-2 Rapid Test | Autobio Diagnostic Co. | Zhengzhou, China |
| Novel Coronavirus 2019 (2019-nCoV) Antibody IgG/IgM Test | Avioq Bio-Tech Co. | Yantai, China |
| Nadal COVID-19 IgG/IgM Test | Nal Von Minden GmbH | Regensburg, Germany |
| Biosynex COVID-19 BSS | Biosynex | Illkirch-Graffenstaden, France |
| 2019-nCoV Ab Test | Innovita Biological Technology Co. | Qian'an, China |
| 2019-nCoV IgG/IgM Test | Biolidics | Mapex, Singapore |
| COVID-19-Check-1 | Veda Lab | Alençon, France |
| Finecare SARS-CoV2 Antibody Test | Guangzhou Wondfo Biotech | Guangzhou, China |
| Wondfo SARS-CoV2 Antibody Test | Guangzhou Wondfo Biotech | Guangzhou, China |

## 2.2 Membrane Extraction using Yolov8

Each image in the dataset required precise annotation to facilitate accurate model training. This process has been performed manually using a specialized annotation tool [34]. The primary region of interest was the membrane area of the rapid test kits, where the test and control lines appear. Annotators drew bounding boxes around these regions, labeling them accordingly. The annotated dataset was then used to train the YOLO v8 neural network [35]. YOLO (You Only Look Once) is a state-of-the-art deep learning model renowned for its speed and accuracy in object detection tasks. The annotated images were preprocessed to ensure compatibility with the YOLO v8 input requirements. This involved resizing the images and normalizing the pixel values. The YOLO v8 architecture was configured to detect the membrane region of the test kits. Transfer learning was used to retrain the pre-existing weights to correctly identify the membranes. Key parameters were optimized through a series of experiments. The model was trained using the preprocessed dataset. The training process involved forward and backward passes through the network, adjusting the weights to minimize the detection error. We employed a standard split of the dataset into training and validation sets, using 80% of the images for training and 20% for validation. The performance of the trained model was evaluated on the validation set. Key metrics such as precision, recall, and mean Average Precision (mAP) were calculated to assess the model's effectiveness in detecting the membrane regions.

### 2.3 Classification of Test Result Kit

The YOLO v8 model is used to detect membrane regions in COVID-19 rapid test kit images, followed by extraction and cropping of these regions from the original images. This preprocessing step ensures that only relevant areas of the test kits are fed into the subsequent classification model, thereby enhancing the accuracy and efficiency of the classification process. A total of 80 batches are generated, with 54 batches allocated for training the Convolutional Neural Network (CNN), and 12 batches each reserved for validation and testing. This allocation results in a distribution of 70% for training, and 15% each for validation and testing, ensuring a balanced representation across the dataset partitions. Prior to input into the CNN model, all images are gray scaled and resized to dimensions of 256 by 256 pixels as shown in Figure 3.

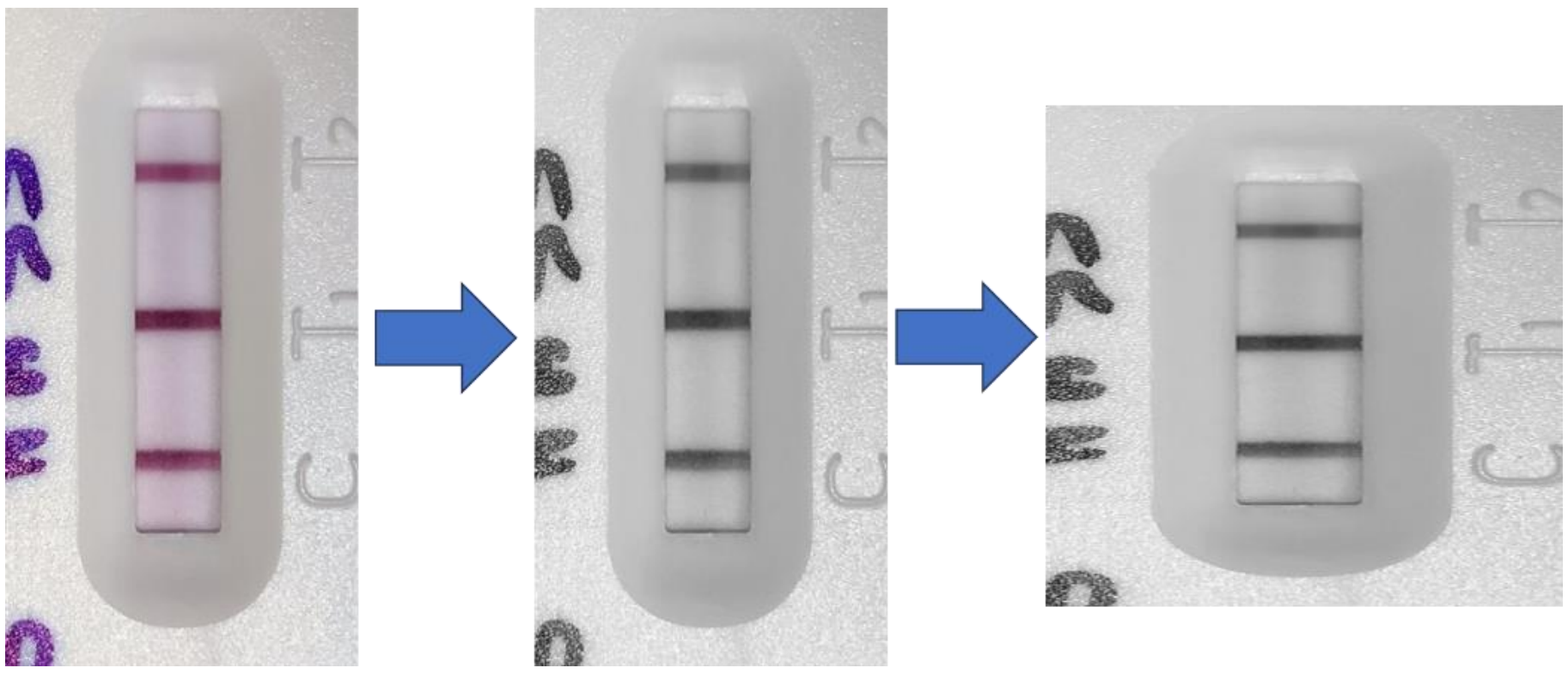

**Figure 3:** Images of the membranes of the test kit are extracted using the YOLO v8 algorithm, the cropped images are then converted into grayscale and subsequently rescaled to 256 × 256 pixel. Image of rapid test kit collected from Mendels et al. [24]

To classify the test results into positive, negative, or invalid categories, a Convolutional Neural Network (CNN) is developed and trained using the cropped membrane images. The CNN architecture is designed to capture the distinctive features of the test lines, enabling accurate classification of the test outcomes. The architecture of the CNN model includes several convolutional layers interspersed with pooling and dropout layers to reduce overfitting. The model's design is illustrated in Figure 4.

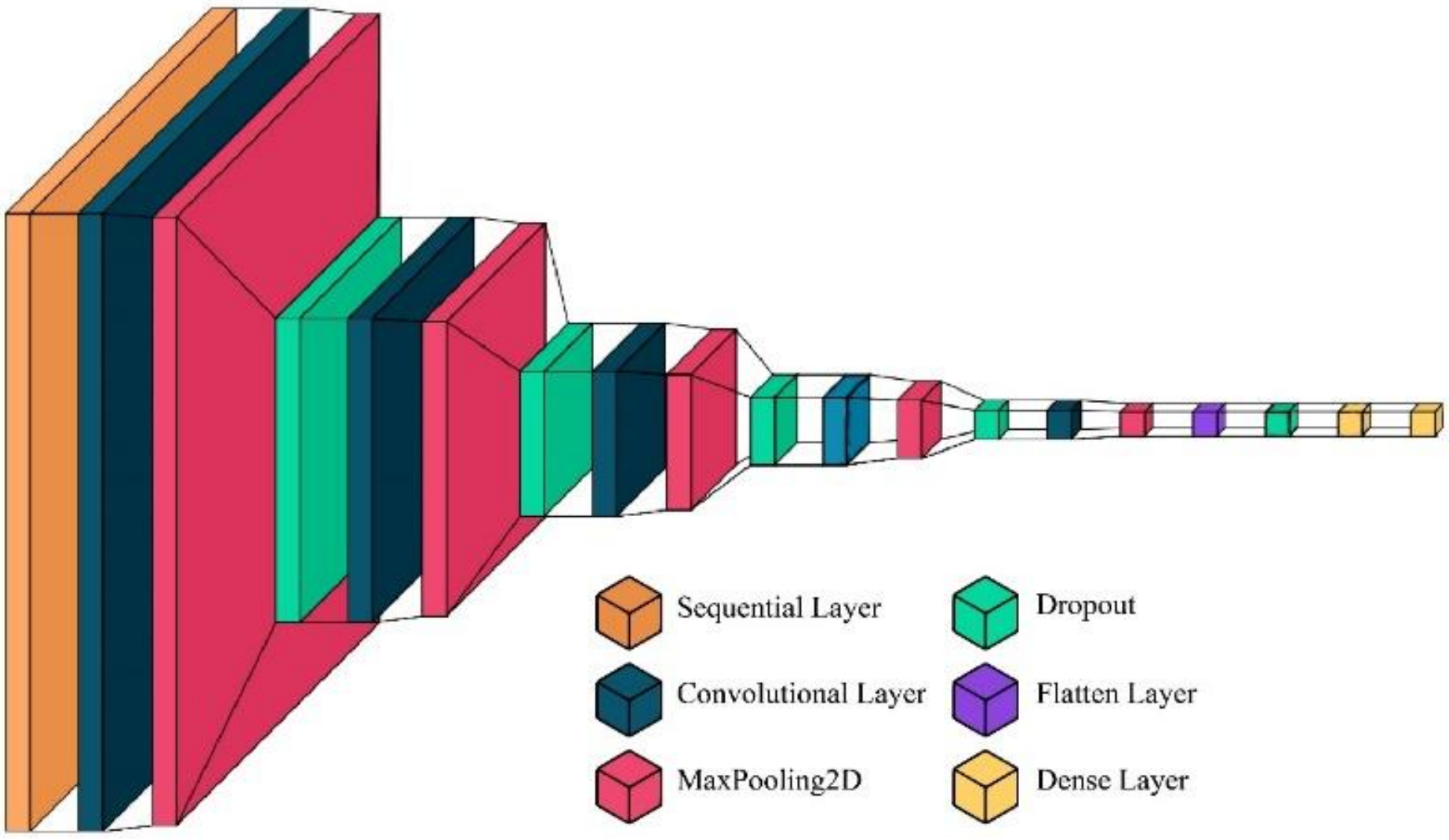

**Figure 4:** Neural network architecture of the Deep Learning Model that has been used in this study where the sequential layer is followed by multiple convolutional layers, max pooling layers, and dropout layers.

As shown in Table 2, the input shape of the sequential model is defined assuming an initial size of 128×128 pixels with 1 channel of grayscale image. This configuration serves as the starting format for subsequent layers in this model. The initial convolutional layer begins with 128 filters of size 3×3 and utilizes Rectified Linear Unit (ReLU) activation to focus on detecting fundamental features such as edges and textures. This layer produces a feature map of size (62, 62, 128), indicating both spatial and depth dimensions of the extracted features. Following the convolutional stage, a max pooling layer employing a 2×2 filter is used to down sample the feature map, reducing its spatial dimensions to (31, 31, 128). This down sampling enhances computational efficiency and translational invariance within the extracted features.

**Table 2:** Representation of our CNN model's architecture

| Layer Name | Output Shape | Activation Function | No. of Filters | Filter Size | Dropout Rate |
|---|---|---|---|---|---|
| Sequential Layer | (32, 128, 128, 1) | - | - | - | - |
| Convolutional Layer (1) | (32, 126, 126, 128) | ReLU | 128 | (3, 3) | - |
| MaxPooling2D (1) | (32, 63, 63, 128) | - | - | (2, 2) | - |
| Dropout (1) | (32, 63, 63, 128) | - | - | - | 0.2 |
| Convolutional Layer (2) | (32, 61, 61, 128) | ReLU | 128 | (3, 3) | - |
| MaxPooling2D (2) | (32, 30, 30, 128) | - | - | (2, 2) | - |
| Dropout (2) | (32, 30, 30, 128) | - | - | - | 0.2 |
| Convolutional Layer (3) | (32, 28, 28, 64) | ReLU | 64 | (3, 3) | - |
| MaxPooling2D (3) | (32, 14, 14, 64) | - | - | (2, 2) | - |
| Dropout (3) | (32, 14, 14, 64) | - | - | - | 0.2 |
| Convolutional Layer (4) | (32, 12, 12, 64) | ReLU | 64 | (3, 3) | - |
| MaxPooling2D (4) | (32, 6, 6, 64) | - | - | (2, 2) | - |
| Dropout (4) | (32, 6, 6, 64) | - | - | - | 0.2 |
| Convolutional Layer (5) | (32, 4, 4, 32) | ReLU | 32 | (3, 3) | - |
| MaxPooling2D (5) | (32, 2, 2, 32) | - | - | (2, 2) | - |
| Flatten | (32, 128) | - | - | - | - |
| Dropout (5) | (32, 128) | - | - | - | 0.2 |
| Dense (1) | (32, 16) | ReLU | - | - | - |
| Dense (2) | (32,3) | Softmax | - | - | - |

The architecture progresses with a second convolutional layer also featuring 128 filters of size 3×3 and ReLU activation. This layer refines the detected features, resulting in an output feature map of (29, 29, 128). Subsequently, another max pooling layer further reduces the spatial dimensions to (14, 14, 128), continuing to enhance computational efficiency while preserving essential feature information. A corresponding dropout layer at 20% rate follows to maintain regularization and prevent overfitting by promoting robust feature learning. The subsequent layers include a third convolutional layer employing 64 filters of size 3×3 and ReLU activation, which extracts increasingly abstract features from the input, resulting in an output feature map of (12, 12, 64). A max pooling layer then reduces the dimensions to (6, 6, 64), benefitting in spatial abstraction and feature localization.

In the fourth layer, 64 filters of size 3×3 are utilized to extract detailed patterns and enhance the network's capability to define internal spatial relationships. An output feature map of (4, 4, 64) is produced, significantly reducing spatial dimensions while preserving feature depth. Subsequent max pooling further reduces dimensions to (2, 2, 64), focusing on critical feature localization and abstraction. A dropout rate of 20% is applied to maintain regularization across deeper layers. The architecture culminates with a fifth convolutional layer where 32 filters of size 3×3 are employed, and ReLU activation is used to capture complex and high-level features essential for final classification decisions. A final max pooling operation is applied, followed by flattening the resulting 2D matrix into a 1D vector. This prepares the data for fully connected layers, starting with a dense layer comprising 16 neurons using ReLU activation to combine and refine learned features from previous layers. The last dense layer of the architecture features neurons equal to the number of classes with Softmax activation, generating output probabilities for each class (positive, negative, or invalid). The class with the maximum output probabilities is selected as the predicted class. The output probability for the predicted class is presented as the confidence level of the prediction. Throughout the entire architecture, dropout layers (dropout rate = 0.2) are strategically employed to ensure robust regularization, mitigate overfitting, and enhance model performance and generalizability across diverse test scenarios and dataset variations. This detailed process of our system is presented in Figure 5 and this process emphasizes the optimization of the model's efficiency and reliability in automated medical diagnostics, particularly critical during public health crises like the COVID-19 pandemic for disabled people.

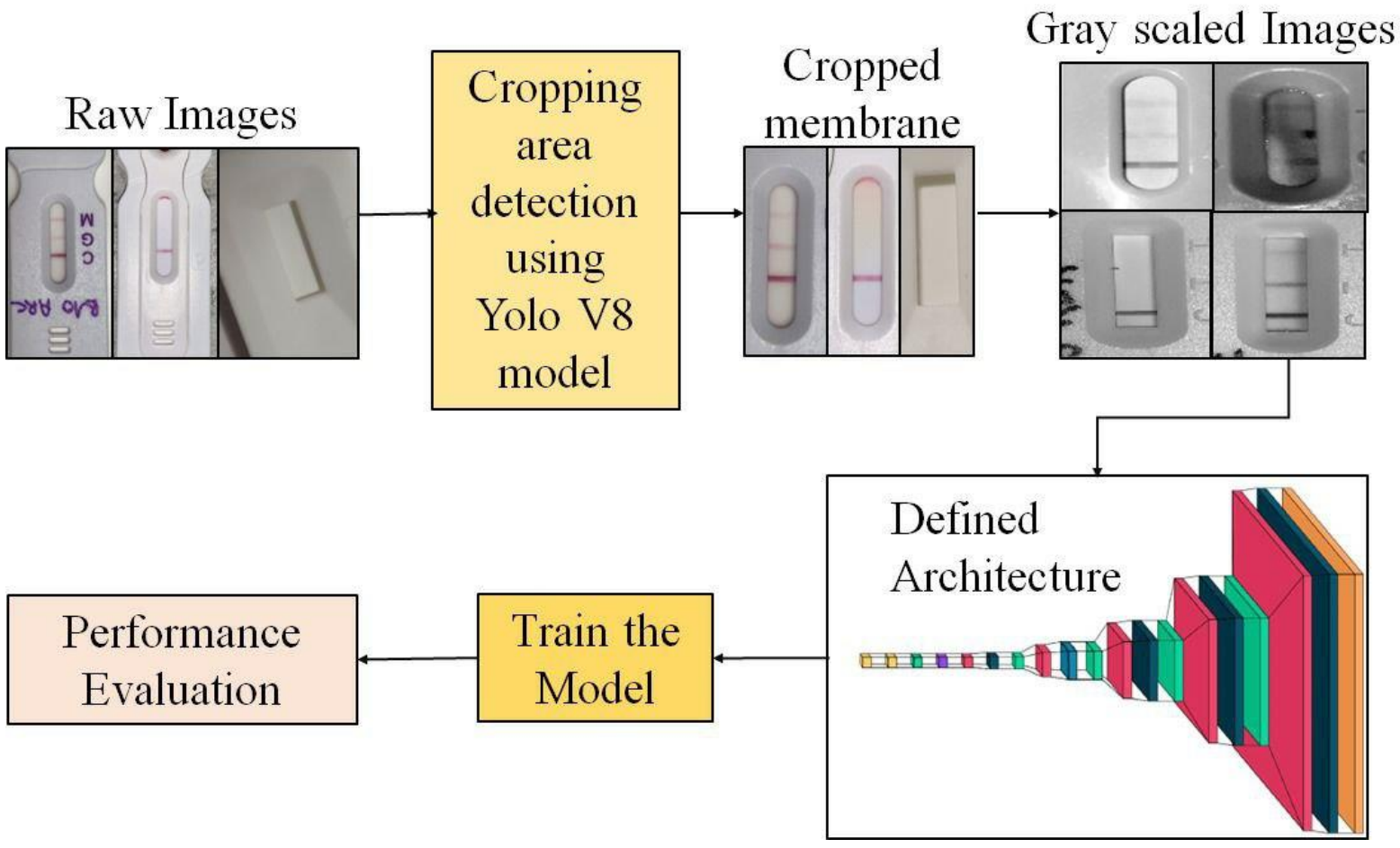

**Figure 5:** The process diagram of our proposed model is shown. First, the raw images are cropped using a custom-trained YOLO v8 model to extract the membrane zones. The cropped images are then converted to grayscale and resized to 128x128 pixels for model training. The model's performance is evaluated based on its accuracy in classifying the test dataset.

### 2.4 SHAP Analysis

To interpret the decisions made by the CNN used for detecting positive, negative, and invalid results in rapid test kits, SHapley Additive exPlanations (SHAP) analysis was employed [36]. SHAP values are based on Shapley values from cooperative game theory and provide a unified measure of feature importance by computing the contribution of each input feature (in this case, pixel values) to the model's output. The DeepExplainer module from the SHAP library, designed for deep learning models, was utilized to compute these values. The analysis began by selecting a representative subset of the training dataset as background data, which is used to generate reference values for calculating the SHAP values. SHAP values were then computed for a sample of test images. These values help us understand how each pixel influences the model's prediction for a specific class (positive, negative, or invalid). Visualizations, including summary plots, dependence plots, and heatmaps, were generated to illustrate the regions of the images that had the most significant impact on the model's decisions.

By examining these visualizations, we can identify which parts of the test kit images the model focuses on to make its predictions. For instance, positive results were associated with high SHAP values in the area where the test line appears, while negative results showed high SHAP values in the control line region and low values in the test line area. Invalid results exhibited high SHAP values in regions indicating anomalies or the absence of expected lines. This analysis confirmed that the model was making decisions based on relevant features and areas of the test kits, providing insights into both correct and incorrect classifications. These insights validate the model's reliability and robustness by ensuring its focus aligns with human-interpretable features.

### 2.5 Mobile app deployment

To automate the analysis of COVID-19 rapid test kits, we have developed a mobile application using Flask as the backend framework [37]. The application uses the YOLO v8 model for membrane extraction and a CNN for classifying the test results as positive, negative, or invalid. The workflow as demonstrated in Figure 6 begins with the user capturing an image of the rapid diagnostic test using their smartphone. This image is then uploaded to Firebase Storage, a robust and scalable cloud storage solution that ensures secure and efficient handling of user-uploaded images. Upon successful upload, Firebase generates a link to the uploaded image, which is sent to the backend server implemented using Flask.

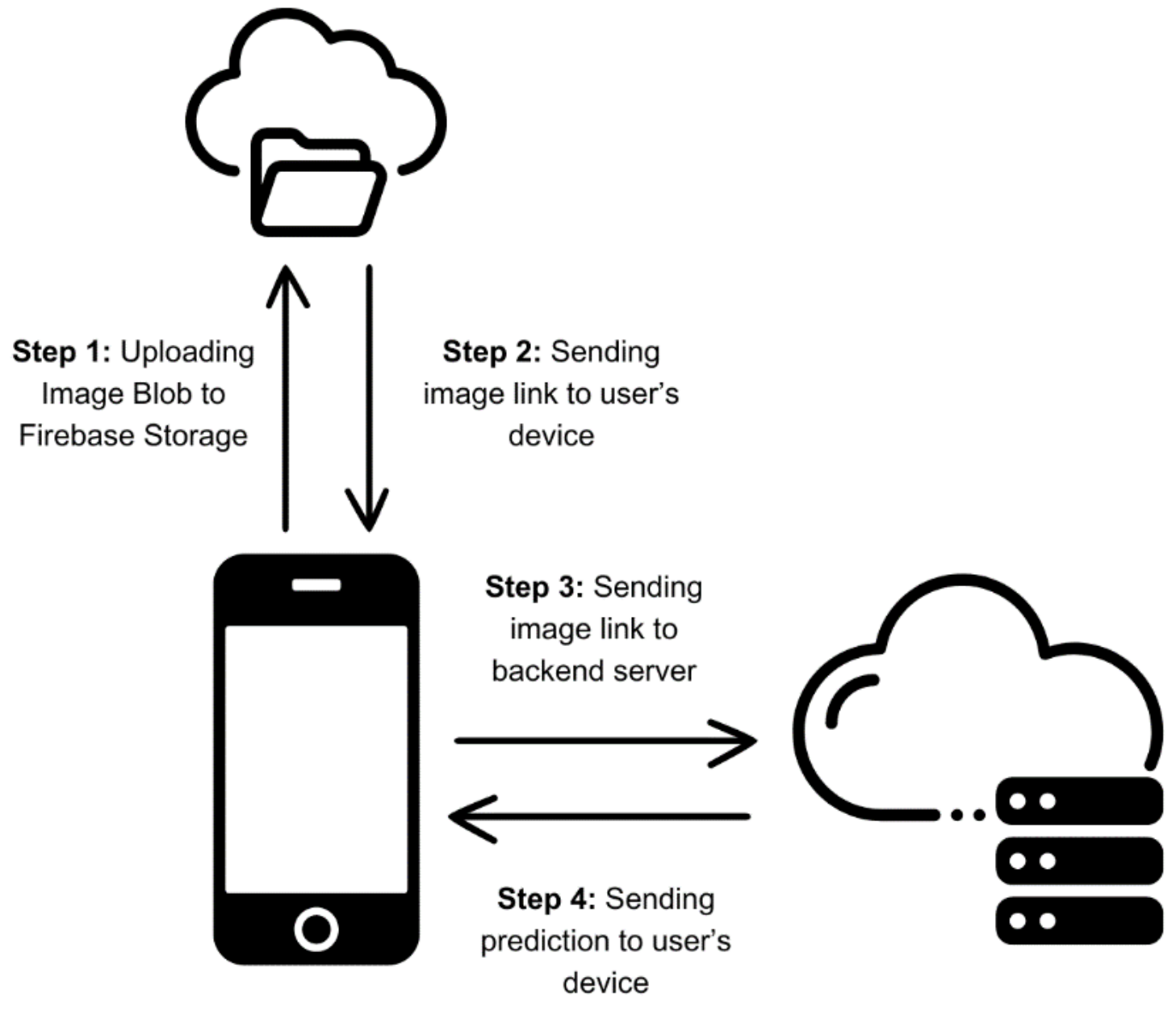

**Figure 6:** Schematic of the working principle of the mobile app developed for automated rapid test kit detection using an Android phone's camera. After an image is taken, the image blob is sent to the Firebase storage, which generates a link to the image. The image link is sent to the backend server which processes the image and generates the prediction before presenting it to the user.

The backend server retrieves the image using the provided link and processes it in two main steps. First, the YOLO v8 model is used to detect and extract the membrane region from the test kit. This extracted region, which contains the critical test lines, is then passed through the pre-trained CNN model to classify the test result. The CNN model outputs the classification as positive, negative, or invalid, along with the associated confidence level. After the analysis is complete, the classification result and its confidence level are sent back to the user's phone. The mobile application then displays the test result, providing an intuitive and user-friendly interface for easy interpretation.

For deployment, Firebase Storage is configured to securely handle image uploads, with appropriate authentication and access rules in place to protect user data. The Flask application is deployed on a cloud server to ensure high availability and scalability. The YOLO v8 and CNN models are integrated into the Flask application, with the models being loaded at server startup to minimize latency during image processing. RESTful APIs are developed to facilitate communication between the mobile application and the backend server, with endpoints for image upload links, processing requests, and result retrieval. The entire system has undergone rigorous testing to ensure accuracy, efficiency, and robustness. performance optimization techniques are applied to minimize response time and enhance the user experience.

## 3 Results and discussion

### 3.1 Results from YoloV8

To accurately extract membrane regions from test kit images, the YOLOv8 model was trained using a set of annotated images. The performance of the model during training and validation was monitored and is depicted in Figure 7.

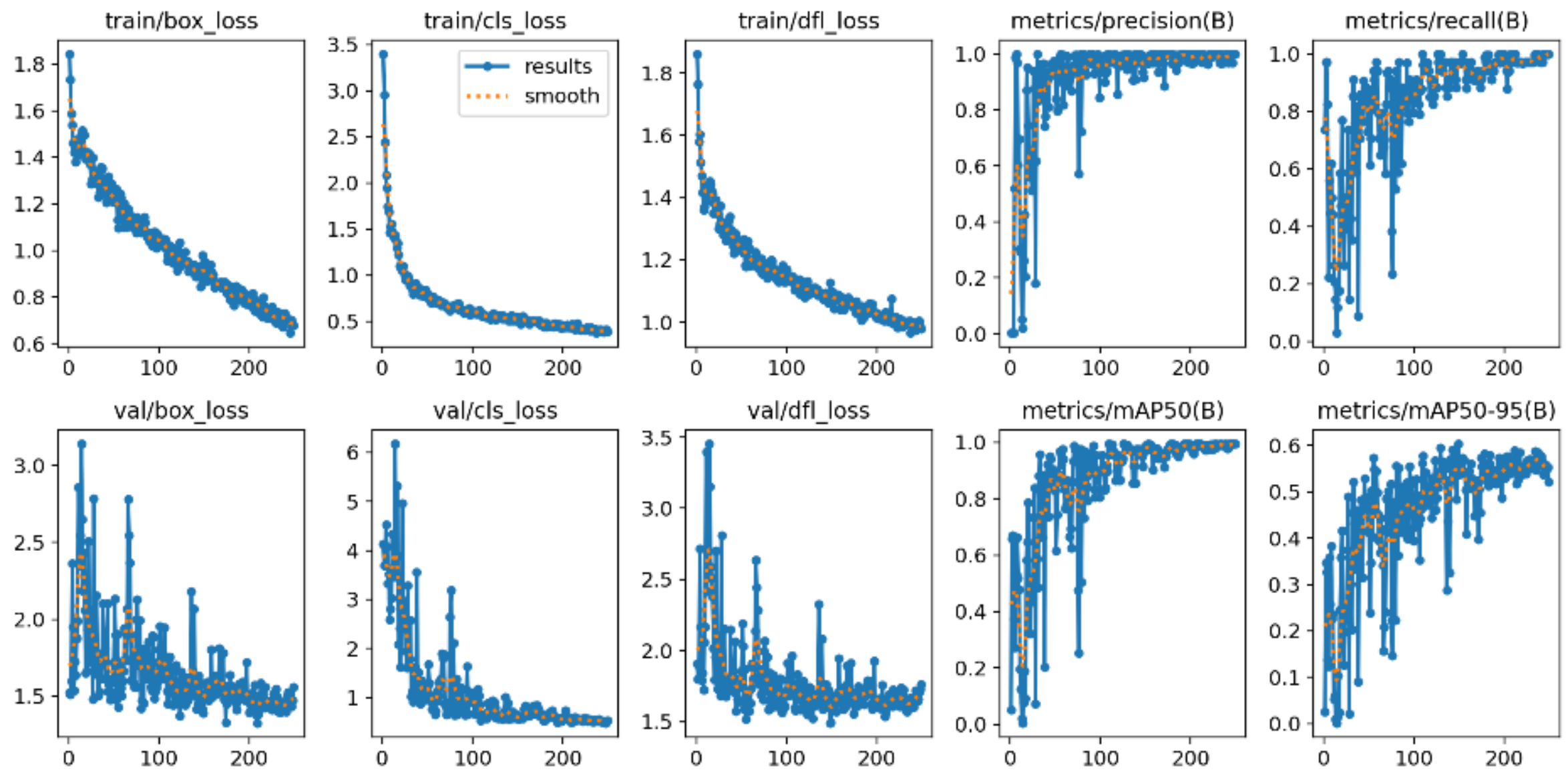

**Figure 7:** Performance Metrics of the YOLO v/8 model implemented to extract and crop out the membrane zone from the images of the rapid test kit. The top row of plots represents the training metrics, including box loss, classification loss, distributional focal loss (DFL), precision, and recall. The bottom row showcases the corresponding validation metrics. Each plot tracks the progression of these metrics over time, with "results" representing the actual metric values and "smooth" indicating the smoothed trend. The loss metrics (box_loss, cls_loss, and dfl_loss) demonstrate a decreasing trend, indicative of the model's improving accuracy. Precision, recall, mean average precision at IoU thresholds of 0.5 (mAP50), and a range from 0.5 to 0.95 (mAP50-95) all show upward trends, suggesting that the model's ability to accurately detect and classify objects is enhanced with more training.

The training loss plots, including box loss, classification loss, and DFL (Distribution Focal Loss), all show a steady decline, indicating that the model is learning effectively and reducing errors as the epochs progress. Specifically, the box loss starts around 1.8 and drops below 0.8, while the classification loss decreases from around 3.5 to below 0.5. Similarly, the DFL loss decreases from approximately 1.8 to around 1.0, further demonstrating the model's improving performance. These downward trends are mirrored in the validation loss plots, though with more variability, suggesting some degree of overfitting or noise in the validation data. Nonetheless, the overall downward trend in validation losses indicates that the model generalizes reasonably well to unseen data. The precision, recall, and mean Average Precision (mAP) metrics provide additional insights into the model's performance. Precision and recall metrics for the validation set exhibit high values, stabilizing close to 1.0, which signifies that the model accurately detects and classifies the membrane regions with few false positives and negatives. The mAP@50 (mean Average Precision at 50% IoU) and mAP@50-95 (mean Average Precision across different IoU thresholds) also demonstrate significant improvements, with mAP@50 reaching close to 1.0 and mAP@50-95 approaching 0.6. These metrics confirm that the YOLOv8 model is highly effective at extracting the relevant membrane regions from the test kit images, maintaining robust performance across different intersection-over-union (IoU) thresholds.

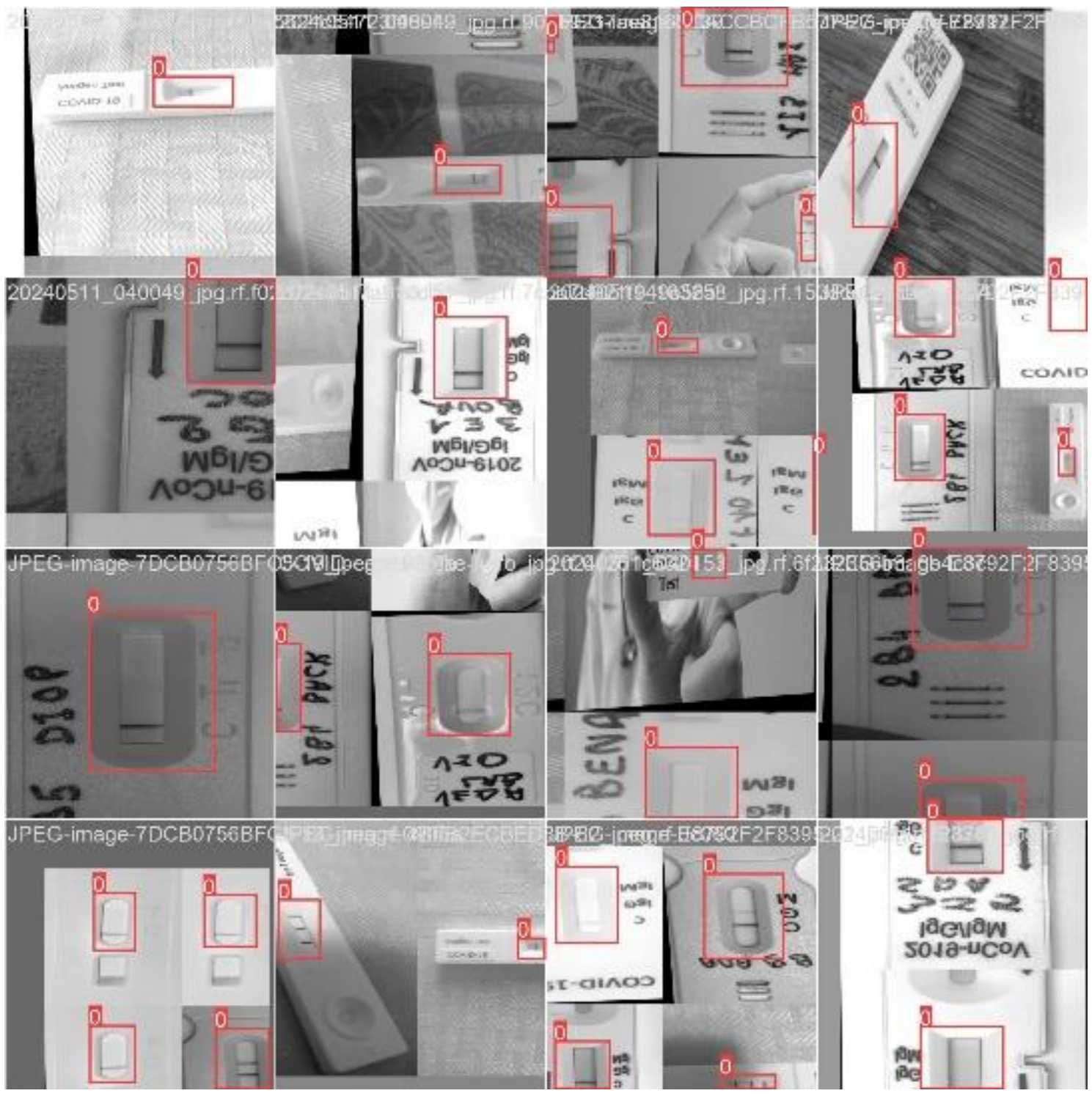

**Figure 8:** YOLO v8 model correctly identifying membranes from images of rapid test kits taken from different angles, lights, and elevations.

Figure 8 showcases YOLOv8 successfully identifying membranes in rapid test kit images taken from various angles and under different lighting conditions. This visual evidence further supports the quantitative metrics, demonstrating the model's robustness and adaptability to diverse real-world scenarios. The model's consistent performance across different environments highlights its practical applicability and reliability in accurately detecting membrane regions, irrespective of image variations.

### 3.2: Results from Classification

The performance of the Convolutional Neural Network (CNN) for classifying rapid antigen testing kit images into positive or negative categories has been evaluated over 200 training epochs. The model's performance metrics, including training and validation accuracies and losses, has been recorded to assess its convergence and generalization capabilities. The accuracy curves for both training and validation sets, as illustrated in Figure 9 (a), indicate a rapid increase in accuracy during the initial epochs. The training accuracy swiftly rose from 50% to above 90% within the first 20 epochs, reaching approximately 99.3% by the 60th epoch. After this point, the training accuracy continued to improve gradually, stabilizing at around 99.3% towards the end of the training process. Similarly, the validation accuracy has exhibited a significant improvement in the early epochs, closely following the training accuracy curve. The validation accuracy initially increased sharply, surpassing 90% within the first 20 epochs, and eventually stabilized at approximately 99.2% by the final epoch. The close alignment of the training and validation accuracy curves suggests that the model is able to generalize well to the validation data, with minimal signs of overfitting. Again, the loss curves for the training and validation sets, as shown in Figure 9 (b), show a marked decrease during the initial epochs. The training loss has rapidly dropped from an initial value of around 1.0 to below 0.1 within the first 20 epochs. By the 50th epoch, the training loss has decreased to approximately 0.05 and continued to decline gradually, stabilizing around 0.018 towards the end of the training process. The validation loss has exhibited a similar trend, decreasing sharply from an initial value of around 1.0 to below 0.1 within the first 20 epochs. After a period of fluctuation between the 20th and 50th epochs, the validation loss has continued to decrease, stabilizing around 0.024 by the final epoch. The parallel reduction in both training and validation losses indicates effective learning by the model and suggests that there is no significant overfitting.

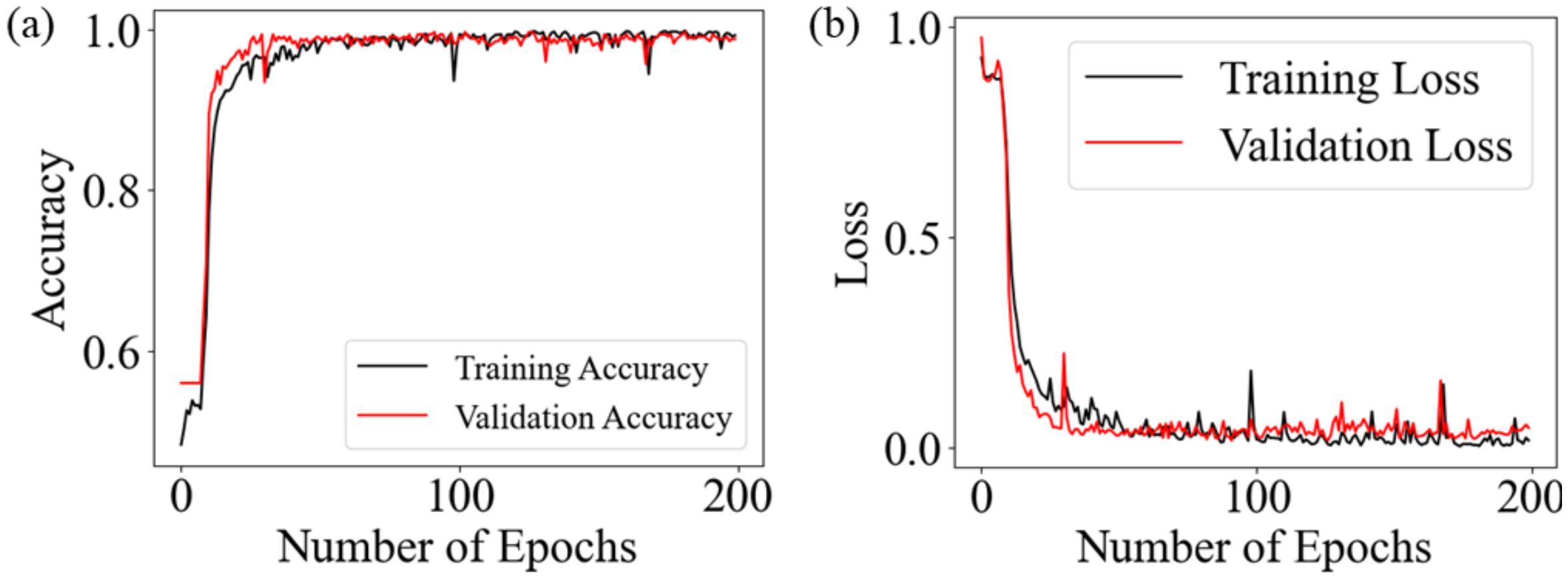

**Figure 9:** Improvement in (a) accuracy and (b) loss with an increasing number of epochs while training the convolutional neural network model.

Figure 10 illustrates the evaluation results on the unseen images, demonstrating the model's capability to accurately predict the class of each image. Each image is accompanied by the actual class, the predicted class, and the confidence score of the prediction. A substantial proportion of the model's predictions perfectly align with the ground truth labels, with a confidence level of 100%. This high confidence reflects the model's strong ability to discern the subtle features of the rapid testing kits, despite variations in shape and structure. While most predictions are made with 100% confidence, a few predictions had confidence scores ranging from 95% to 99%. These slightly lower confidence scores indicate instances where the model was nearly certain but exhibited minimal uncertainty. The model demonstrated consistent performance across different classes (Positive, Negative, Invalid). This consistency is indicative of the model's balanced learning and its ability to generalize well across different types of rapid testing kits. A detailed visual comparison among the first, second, third and fourth row can be seen. First row displays images of rapid testing kits predicted as Positive with 100% confidence, matching the actual class. Second row contains images predicted as Negative with 100% confidence, accurately reflecting the actual class. Third row includes images predicted as Invalid with 100% confidence, corresponding correctly with the actual class. Fourth row shows images with varying confidence levels between 95% and 100%, including one instance of a Positive prediction with 96.25% confidence. Despite the slightly lower confidence, the predictions are still correct.

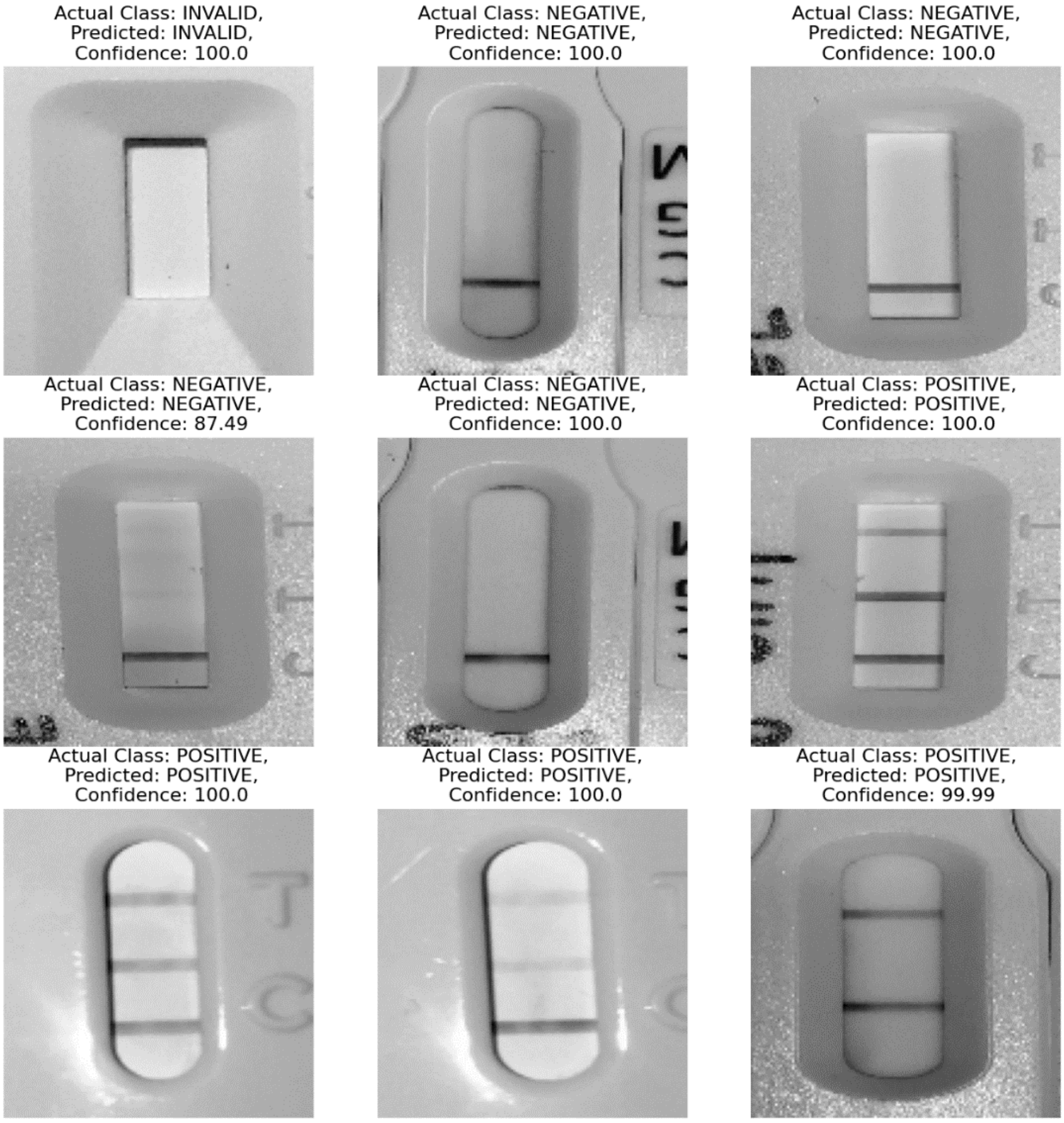

**Figure 10:** Sample predictions by the model presented in this study on test data along with prediction confidence. All three classes, "POSITIVE", "NEGATIVE" and "INVALID" are seen to be correctly classified by the model. This Figure includes images from the work of Mendels et al. [24]

The confusion matrix in Figure 11 illustrates the performance of our classification model across three categories: Invalid, Negative, and Positive. The matrix reveals that the model correctly identified 30 Invalid instances, 222 Negative instances, and 200 Positive instances, with minimal misclassification. Specifically, the model misclassified one Negative instance as Invalid, one Negative instance as Positive, and two Positive instances as Negative. This results in False Positives (FP) of 1 for Invalid, 2 for Negative, and 1 for Positive, while the False Negatives (FN) are 0 for Invalid, 1 for Negative, and 2 for Positive. The True Negatives (TN), which are the correctly identified non-class instances, are 424 for Invalid, 230 for Negative, and 253 for Positive. For the Invalid class, the model achieved a precision of 0.97, a recall of 1.00, and an F1-score of 0.985, indicating a high ability to correctly identify Invalid instances with few False Positives. The Negative class showed a precision of 0.99, a recall of 0.995, and an F1-score of 0.993, suggesting that the model is highly effective at identifying Negative instances with minimal misclassification. Similarly, the Positive class exhibited a precision of 0.995, a recall of 0.990, and an F1-score of 0.992, demonstrating the model's strong performance in accurately detecting Positive instances with few errors. Thus, the model achieved an impressive accuracy of approximately 99.1%, highlighting its robustness in classification tasks. The macro-averaged precision, recall and F1-score are 0.985, 0.995, and 0.990, respectively.

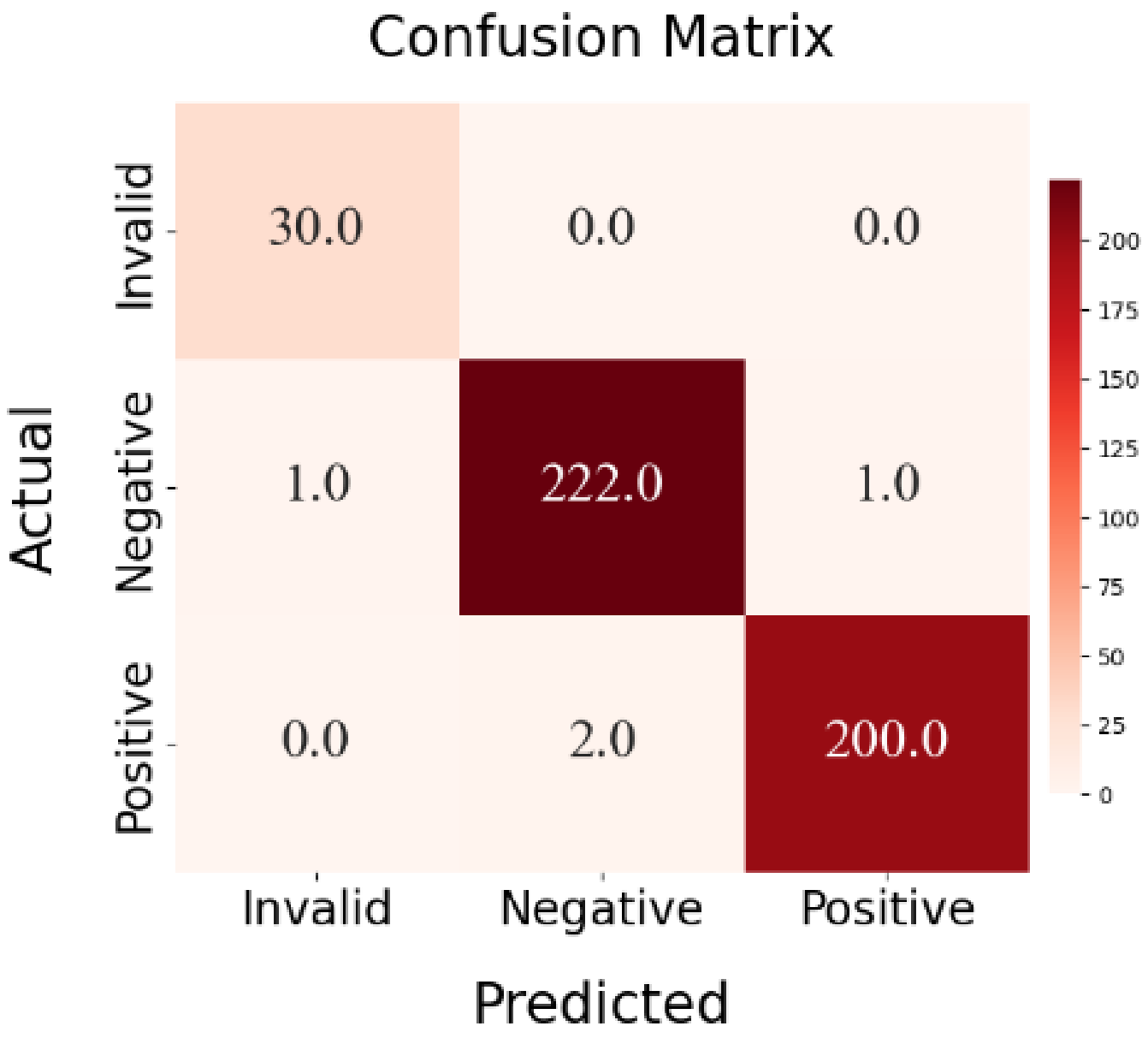

**Figure 11:** Confusion matrix of the predictions on test data by the CNN model is presented where an accuracy of more than 99% is observed.

### 3.3 SHAP Analysis

A critical aspect of this study involves discerning the factors contributing to erroneous classifications of positive, negative, or invalid results by the model across various images. We conducted a SHAP analysis to gain insights into the model's decision-making process for classifying pregnancy test results. SHAP analysis assigns attribution scores to different features in the image that contribute to the final prediction. A positive attribution score from a feature indicates the feature supported the model's classification, while a negative score indicates the feature opposed the classification. By interpreting the SHAP scores for various test images, we can gain insights into the factors that the model relied on to make predictions. This can help identify if the model is using relevant visual cues from the pregnancy test membrane for classification and also highlight potential biases or weaknesses in the model's decision process.

We first have a look at Figure 12 containing the samples that the model correctly classifies to see if it is making its decisions based on the correct cues. We can clearly see the model is making its decisions based on the test and control lines where the presence or absence of those lines heavily influences their decision. Each row represents different test images, while the 2nd column corresponds to the classification outcomes. SHAP values are visualized with red areas indicating a positive influence and blue areas indicating a negative influence on the model's decisions. For positive classifications, the model focuses on regions where the test line appears, as highlighted by the red areas in the "Positive" column. These regions, with high SHAP values, significantly contribute to the model identifying a test as positive.

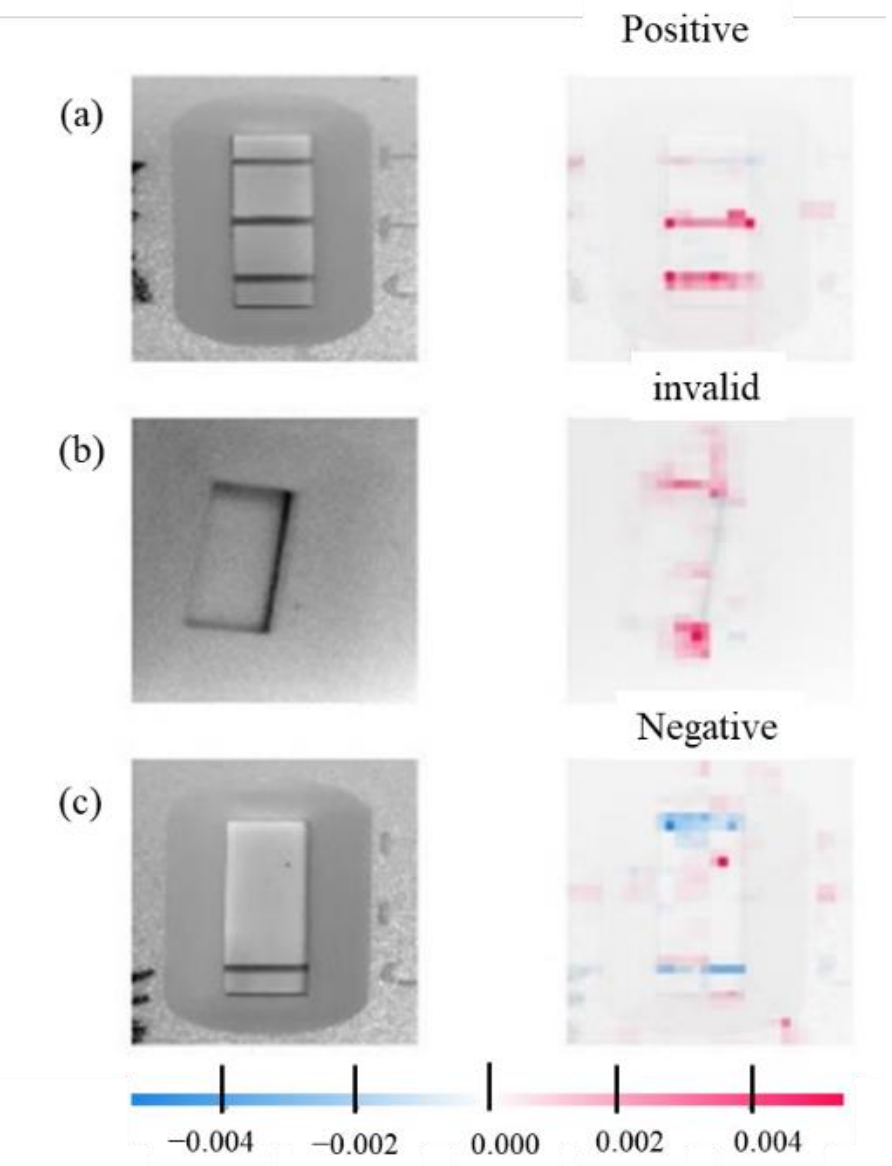

**Figure 12:** SHAP Analysis of three sample correctly classified samples from the test dataset. a) A Positive Sample Predicted as Positive b) An Invalid sample predicted as Invalid c) A Negative sample predicted as Negative.

The consistent presence of red highlights across different images underlines the model's reliability in detecting the test line, even under varied conditions such as different lighting and angles. In the case of negative classifications, the SHAP analysis shows red areas in the "Negative" column, focusing on the regions where the control line is present. For invalid results, the model highlights irregularities or the absence of expected lines, as seen by the red and blue areas in the "Invalid" column. This indicates the model's capability to detect anomalies or issues with the test kits, ensuring a robust and realistic approach to rapid test detection. The varied images and consistent SHAP value patterns demonstrate the model's adaptability and accuracy across different conditions.

Figure 13 shows instances where the model erroneously categorized test kit images. Out of 456 test kits, there were only four classification errors. External marks, smudges, stains, or ink marks on the test kits seem to be the main factors leading to these misclassifications.

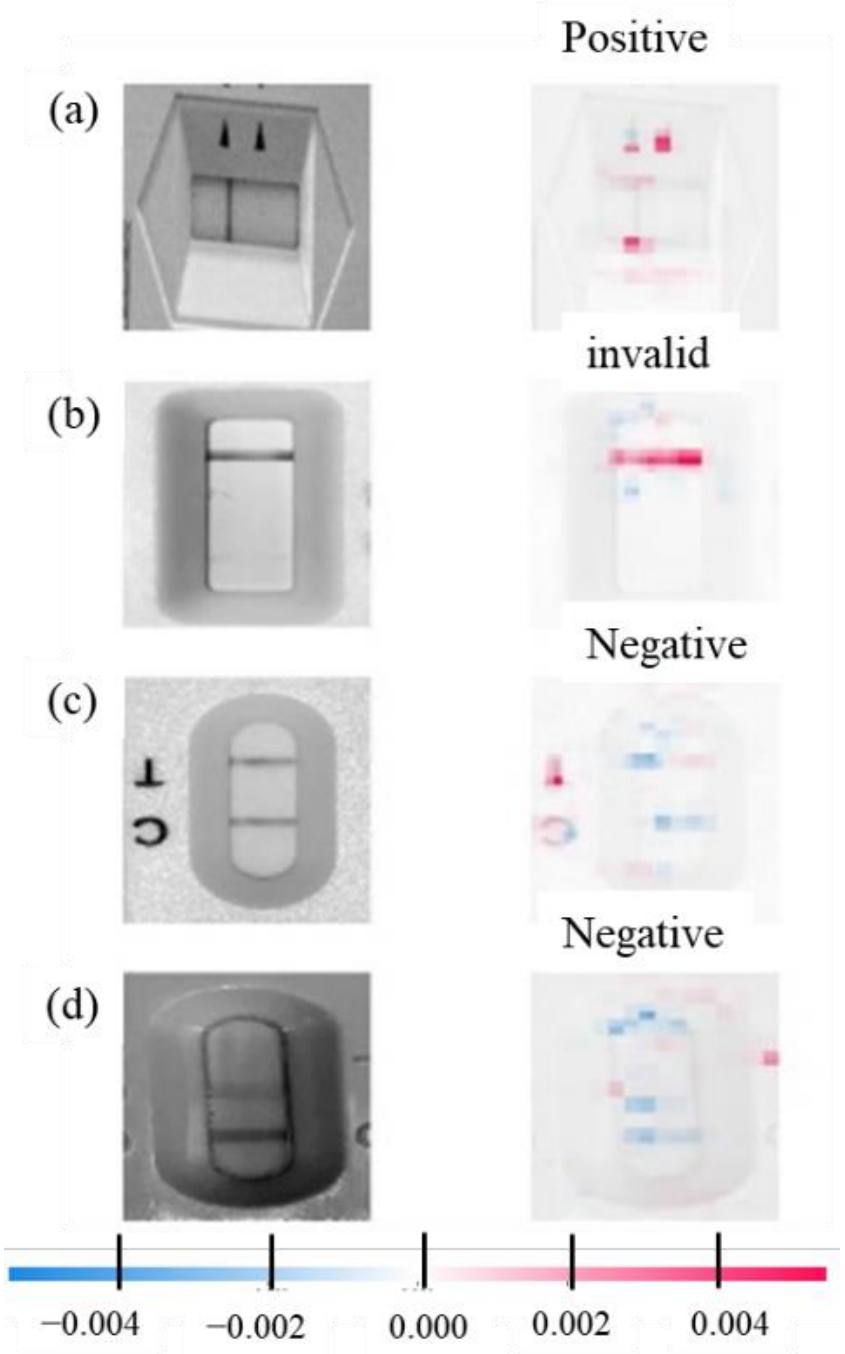

**Figure 13:** SHAP Analysis of all four incorrect results from the predictions on test dataset a) A Negative sample classified as Positive b) A Negative Sample classified as Invalid c) A Positive sample classified as Negative d) A Positive sample classified as Negative.

In Figure 12(a), the sample is incorrectly classified as Positive instead of Negative, largely due to the arrow markings on the test kit, which are highlighted in red just outside the membrane zone. In Figure 12(b), the model predicted the test kit as "Invalid" because of the faded control line, indicated by dark red markings. Figure 12(c) shows a positive test kit wrongly classified as Negative, with blue markings on the test line indicating the model's lack of confidence. Lastly, in Figure 12(d), the test line is very faint, causing the model to predict it as Negative, but the blue markings on the faded line show that those pixels opposed the Negative classification.

### 3.4 Mobile App Implementation

The mobile app that was developed shows great accuracy and provides a reasonably fast prediction. After the user takes a photo, it takes 6 seconds (depending on the user's internet speed) to be uploaded to the server and 5 seconds to deliver a prediction. Within 11 seconds a user is able to receive the result of the rapid test kit via the user-friendly app.

Figure 14 illustrates the usage of an app designed for detecting COVID-19 rapid test results. In Figure 14(a), the user captures a photo of the rapid test kit using the app's camera function. The captured image shows a single line in the test area, indicating a potential test result. Figures 14(b) and 14(c) display the app's process of uploading and analyzing the photo. The "Uploading Photo." message signifies the image is being sent to the server or cloud for processing, while the "Analyzing Photo." message indicates the app is performing image analysis to determine the test result. Figure `14(d) presents the final result, with the app concluding that the test result is "NEGATIVE" with a confidence level of 99.99%. This step-by-step process showcases the app's ability to automate the analysis of rapid test kits, providing users with immediate feedback. The high confidence level suggests a robust underlying model, capable of accurately interpreting the test results. This application demonstrates significant potential in simplifying and speeding up the process of reading rapid test kits, reducing the likelihood of human error, and improving accessibility for users conducting tests at home.

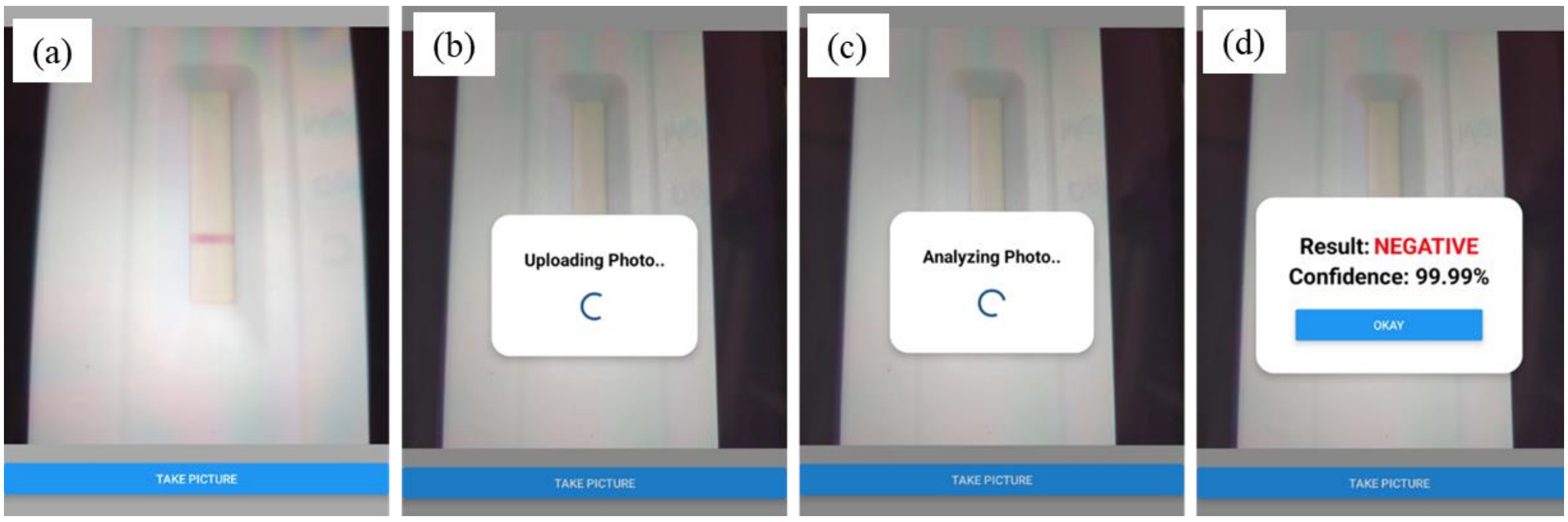

**Figure 14:** Use case for the application developed to assist in rapid testing kit inspection. (a) Capturing the image of a rapid test kit (b) Uploading the photo to a server (c) Analyzing the image (d) Providing the user with the prediction.

## 4. Conclusions

This study presents an innovative approach to enhancing the accuracy, accessibility, and reliability of rapid test result interpretation by integrating smartphone-aided AI, making it more accessible for visually impaired users. By leveraging AI algorithms, including YOLOv8 and CNN, the system allows users, even those with visual impairments, to capture test images without needing perfect alignment, promoting greater independence. YOLOv8 accurately detects and crops the membrane region, even when the test kit is off-center or at the image's edge. This improves sensitivity and specificity by enabling the CNN model to focus solely on the membrane zone, allowing it to better detect faint lines and differentiate genuine test lines from background noise. The YOLOv8 model exhibited significant reductions in training losses—box loss decreased from 1.8 to below 0.8, classification loss from 3.5 to below 0.5, and Distribution Focal Loss (DFL) from 1.8 to around 1.0—indicating effective learning and error reduction. Despite some variability, the validation loss plots mirrored the downward trend seen in training losses. The mAP@50 reached close to 1.0 and mAP@50-95 approached 0.6, suggesting the model generalizes well to unseen data. Precision and recall for the validation set stabilized close to 1.0, confirming the YOLOv8 model's robustness and accuracy in detecting and classifying membrane regions with minimal false positives and negatives. The CNN model consistently predicted test results with high confidence, achieving around 99.1% accuracy. The confusion matrix showed precision, recall, and F1-scores of 0.97, 0.99, and 0.995 for invalid, negative, and positive classes, respectively. SHAP analysis confirmed the model's reliance on relevant visual cues, such as test and control lines, and identified external marks, smudges, and faded regions as primary factors for misclassification. The developed mobile application demonstrated high accuracy and quick prediction times, delivering results within 11 seconds—6 seconds for image upload and 5 seconds for prediction—highlighting its practical applicability. The study underscores the transformative potential of integrating AI with smartphone technology for rapid test diagnostics, particularly by enhancing accessibility for visually impaired users. Future versions of the mobile application could incorporate interactive interfaces that guide users through optimal image capture, offering real-time feedback on test alignment and lighting conditions to further reduce the risk of errors. Integrating the app with other health monitoring systems, such as telemedicine platforms or wearable health devices, could create a more comprehensive diagnostic solution, offering users a seamless way to track their health status over time.